\documentclass[10pt,twocolumn,letterpaper]{article}

\usepackage{cvpr}              

\usepackage{graphicx}
\usepackage{amsmath}
\usepackage{amssymb}
\usepackage{booktabs}
\usepackage{tabularx}
\usepackage{amsmath}
\usepackage{enumitem}
\usepackage{bbm}
\usepackage{upgreek} 
\usepackage{multirow}
\usepackage{colortbl}
\usepackage{pifont}%
\usepackage{mathtools}

\usepackage[accsupp]{axessibility}  

\newcommand{\hjk}[1]{{\color{black}#1}}

\newcommand{\jhprev}[1]{{\color{black}#1}}

\usepackage[pagebackref,breaklinks,colorlinks]{hyperref}

\usepackage[capitalize]{cleveref}
\crefname{section}{Sec.}{Secs.}
\Crefname{section}{Section}{Sections}
\Crefname{table}{Table}{Tables}
\crefname{table}{Tab.}{Tabs.}

\def\cvprPaperID{11620} 
\def\confName{CVPR}
\def\confYear{2022}
\definecolor{Gray}{gray}{0.85}

\begin{document}

\title{Self-Supervised Path Consistency Learning for HOI Detection }
\title{DP-Aug: Decoding Path Augmentation for Transformers in HOI Detection}
\title{Consistency Learning via Decoding Path Augmentation \\for Transformers in Human Object Interaction Detection}

\author{Jihwan Park$^{1,2}$\quad
SeungJun Lee$^1$ \quad
Hwan Heo$^1$ \quad
Hyeong Kyu Choi$^1$ \quad 
Hyunwoo J. Kim$^{1,}$\thanks{corresponding author.}\\
{\normalsize $^1$Department of Computer Science and Engineering, Korea University \quad
$^2$Kakao Brain} \\
{\tt\small \{jseven7071, lapal0413, gjghks950, imhgchoi, hyunwoojkim\}@korea.ac.kr}\\
{\tt\small \{jwan.park\}@kakaobrain.com}
} 



\newcommand{\posE}{\text{p}}
\newcommand{\R}{\mathbb{R}}

\newcommand{\enc}{\textbf{enc}}
\newcommand{\dec}{\textbf{dec}}

\newcommand{\paths}{\mathcal{P}}

\newcommand{\perm}{\sigma}
\newcommand{\perminv}{\widetilde{\sigma}}
\newcommand{\feature}{\boldsymbol{F}}

\newcommand{\comb}[2]{{}_{#1}\mathrm{C}_{#2}}

\newcommand{\xmark}{\ding{55}}
\newcommand{\scone}{{AP}${}^{}_{\text{role1}}$}
\newcommand{\sctwo}{{AP}${}^{}_{\text{role2}}$}

\maketitle
\begin{abstract}
Human-Object Interaction detection is a holistic visual recognition task that entails object detection as well as interaction classification.
Previous works of HOI detection has been addressed by the various compositions of subset predictions, \eg, Image $\rightarrow$ HO $\rightarrow$ I, Image $\rightarrow$ HI $\rightarrow$ O.
Recently, transformer based architecture for HOI has emerged, which directly predicts the HOI triplets in an end-to-end fashion (Image $\rightarrow$ HOI).  
Motivated by various inference paths for HOI detection, 
we propose cross-path consistency learning (CPC), which is a novel end-to-end learning strategy to improve HOI detection for transformers by leveraging augmented decoding paths.
CPC learning enforces all the possible predictions from permuted inference sequences to be consistent. 
This simple scheme makes the model learn consistent representations, thereby improving generalization without increasing model capacity.
Our experiments demonstrate the effectiveness of our method, and we achieved significant improvement on V-COCO and HICO-DET compared to the baseline models. 
Our code is available at \href{https://github.com/mlvlab/CPChoi}{https://github.com/mlvlab/CPChoi}.
\vspace{-3mm}
\end{abstract}

\label{sec:intro}
\section{Introduction}

Human-Object Interaction (HOI) detection is a holistic visual recognition task that includes detecting individual objects as \texttt{<human, object>}, while properly classifying the type of \texttt{<interaction>}. 
Previous HOI detectors~\cite{gkioxari2018detecting,wang2020learning,li2020pastanet,ulutan2020vsgnet} were mainly built on object detection models. 
They commonly extend CNN-based object detectors~\cite{ren2016faster,lin2017feature,newell2016stacked} with an additional head for interaction classification, \eg, humans and objects are detected first, and their interaction is associated subsequently.

To alleviate the high computation cost of such two-stage HOI detection methods, one-stage models~\cite{wang2020learning,Kim2020UnionDet,liao2019ppdm} have been proposed for faster detection.
These models perform interaction prediction and object detection in parallel.
They compensate for their lower performance with auxiliary predictions for the HOI subsets, \ie, auxiliary predictions for subset \texttt{<human, interaction>} or \texttt{<object, interaction>} may help HOI prediction through post-processing.
However, these works demand different network architectures for each auxiliary prediction, due to strict disciplines for each network's input.
Hence, to introduce flexibility, transformer-based architectures~\cite{kim2021hotr, tamura2021qpic, chen2021reformulating, touvron2021training} have recently been adopted for HOI detection. 
They reformulate the HOI detection problem as a direct set prediction building on  DETR~\cite{carion2020end}.

\begin{figure}[tp]
\begin{subfigure}[t]{.32\linewidth}
\includegraphics[width=\textwidth]{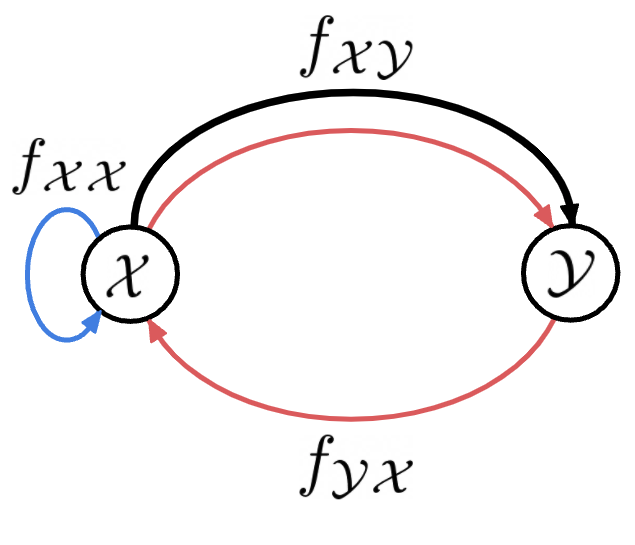}
\caption{cycle consistency}
\end{subfigure}
\begin{subfigure}[t]{.32\linewidth}
\includegraphics[width=\textwidth]{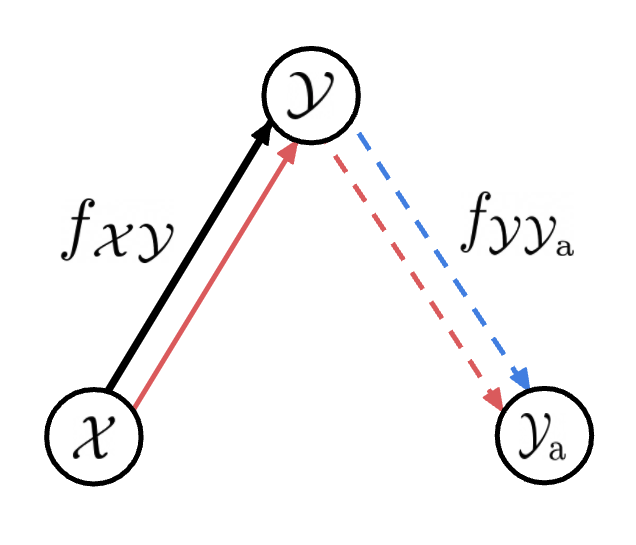}
\caption{cross-task consistency}
\end{subfigure}
\begin{subfigure}[t]{.32\linewidth}
\includegraphics[width=\textwidth]{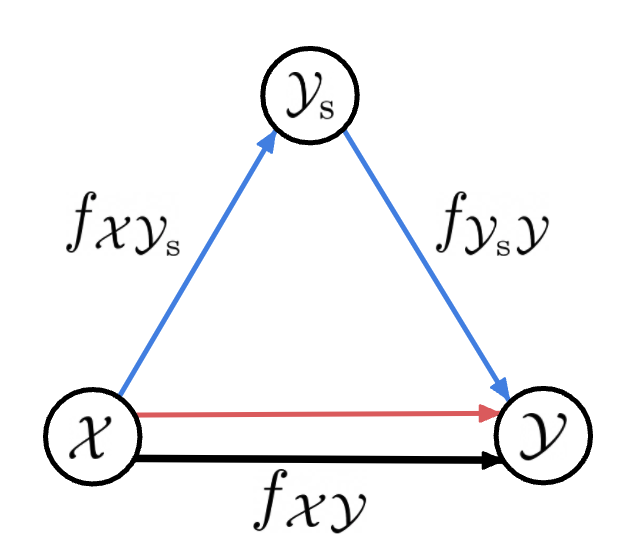}
\caption{cross-path consistency}
\end{subfigure}
\caption{\textbf{Comparison on the variants of consistencies}. The black line refers to the main task function $f_{\mathcal{X}\mathcal{Y}}$, and the {\color{red}red}, {\color{blue}blue} lines refer to the pair of tasks trained to be consistent with each other. (a) Cycle consistency enforces the composite function of $f_{\mathcal{Y}\mathcal{X}} \circ f_{\mathcal{X}\mathcal{Y}}$ to be consistent with identity function $f_{\mathcal{X}\mathcal{X}}$. (b) Cross-task consistency requires an auxiliary pretrained network $f_{\mathcal{Y}\mathcal{Y_\text{a}}}$, represented in dashed lines, to give consistent outputs across tasks. (c) Cross-path consistency does not require task-specific pretrained networks. The output of main task function $f_{\mathcal{X}\mathcal{Y}}$ should be consistent with the composition of the outputs from sub-task functions $f_{\mathcal{X}\mathcal{Y_\text{s}}}$ and $f_{\mathcal{Y}_\text{s}\mathcal{Y}} \circ f_{\mathcal{X}\mathcal{Y}_\text{s}}$.}
\label{fig:consistency variants}
\end{figure}

Motivated by various inference paths in HOI detectors, we propose a simple yet effective method to train HOI transformers. 
We augment the decoding paths with respect to the possible prediction sequences of HOI triplets.  
Then, with the cascade structure of transformers, an input query is sequentially decoded into auxiliary sub-task outputs and the final output.
The stage of each augmented paths stage shares a decoder, in a multi-task learning fashion. 
We further improve our method to leverage the augmented decoding paths by enforcing the outputs from the various paths to be consistent. 
Accordingly, we propose \textbf{Cross-Path Consistency (CPC) Learning}, which aims to predict HOI triplets regardless of inference sequences.


Similar to cross-\textit{task} consistency~\cite{zamir2020robust}, cross-\textit{path} consistency retains \textit{inference path invariance}. 
However, cross-path consistency learning does not require additional pre-trained networks. 
In contrast to cross-task consistency, which demands an auxiliary network to train the main task $\mathcal{X} \rightarrow \mathcal{Y}$ (Figure~\ref{fig:consistency variants}-(b)), cross-path consistency defines an auxiliary domain $\mathcal{Y}_s$ in between $\mathcal{X}$ and $\mathcal{Y}$ (Figure~\ref{fig:consistency variants}-(c)).
In other words, the main task $\mathcal{X} \rightarrow \mathcal{Y}$ (\ie, Image$\rightarrow$HOI) is divided into subtasks $\mathcal{X} \rightarrow \mathcal{Y}_s$ and $\mathcal{Y}_s \rightarrow \mathcal{Y}$ (\eg, Image$\rightarrow$HO$\rightarrow$I).
The main task function $f_{\mathcal{X}\mathcal{Y}}$ is then trained by enforcing its output and the composition of sub-task predictions to be consistent.
Moreover, cross-path consistency learning is temporarily adopted for training only.

Our training strategy can be generalized to any transformer based architecture, and can be applied in an end-to-end method.
Extensive experiments show that HOI transformers trained with CPC learning strategy achieves substantial improvements in two popular HOI detection benchmarks: V-COCO and HICO-DET. 
The contribution of this work can be \jhprev{summarized as the followings}:
\begin{itemize}
    \jhprev{
    \item We propose \textbf{Cross-Path Consistency} (CPC) learning, which is a novel end-to-end learning strategy to improve transformers for HOI detection leveraging various inference paths. 
    In this learning scheme, we use \textbf{Decoding-Path Augmentation} to generate various inference paths which are compositions of subtasks with a shared decoder for effective training.  
    }
    \item Our training scheme achieves substantial improvements on V-COCO and HICO-DET without increasing \textit{model capacity} and \textit{inference time}.  
\end{itemize}







\section{Related Works}
\subsection{Human Object Interaction Detection}
\label{section2.1}
Human-Object Interaction (HOI) detection has been proposed in ~\cite{gupta2015visual}.
Later, human-object detectors have been improved using human or instance appearance and their spatial relationship~\cite{gkioxari2018detecting, kolesnikov2018detecting, gao2018ican}.
On the other hand, graph-based approaches~\cite{qi2018learning, ulutan2020vsgnet, Gao2020DRG, Wang2020Hetero} have been proposed to clarify the action between the \texttt{<human,  object>} pair.

HOI detection models based on only visual cues often suffer from the lack of contextual information. 
Thus, recent works utilize external knowledge to improve the quality of HOI detection.
Human pose information extracted from external models~\cite{cao2017realtime,chen2018cascaded,he2017mask, li2019crowdpose} or linguistic priors and knowledge graph models show meaningful improvement in performance~\cite{zhou2019relation, Liu2020FCMNet, peyre2018detecting,xu2019learning,gupta2017aligned,gkanatsios2019deeply,li2020pastanet, Xubin2020PDNet, Liu2020ConsNet}.

Since the majority of the previous works are based on two-stage methods with slower inference time, attempts for faster HOI detection by introducing simple end-to-end multi-layer perceptrons~\cite{gupta2019no}, or directly detecting interaction points~\cite{wang2020learning,liao2019ppdm}, or union regions~\cite{Kim2020UnionDet, Hou2020VCL, yonglu2020idn} have been suggested. 
\subsection{Transformers in Computer Vision}
\label{section2.2}
Transformer has become the state-of-the-art method in many computer vision tasks. 
In image classification, ~\cite{dosovitskiy2020image} has shown competitive performance on ImageNet without any convolution layers.
DeiT~\cite{touvron2021training} applied knowledge distillation to data-efficiently train the vision transformer.
To extract multi-scale image features, Swin Transformer~\cite{Liu_2021_ICCV} proposed shifted window based self-attention modules that effectively aggregate small patches to increase the receptive field.
In the object detection task, DETR~\cite{carion2020end} has proposed an end-to-end framework eliminating the need for hand-designed components.
DETR's bipartite matching loss between the predicted set and the ground truth labels enables direct set prediction at inference.
Recently, DETR's late convergence problem has been tackled in~\cite{zhu2020deformable,meng2021conditional,Gao_2021_ICCV}.

Inspired by DETR, transformer-based  HOI (Human-Object Interaction)  detectors~\cite{kim2021hotr,tamura2021qpic,chen2021reformulating,zou2021end,dong2021visual} have been recently proposed. 
HOI transformer models have two types of structure, one decoder model and the two decoder model.
The one-decoder model which follows the structure of DETR~\cite{carion2020end} predicts  triplets from the output of a single decoder. 
QPIC~\cite{tamura2021qpic} and HoiT~\cite{zou2021end} are one-decoder models that output \texttt{<human, object, interaction>} triplets directly with multiple interaction detection heads. 
Two-decoder models use two transformer decoders to output distinctive targets. 
For instance, HOTR~\cite{kim2021hotr} and AS-NET~\cite{chen2021reformulating} are composed of an instance decoder that outputs object and an interaction decoder that outputs interaction. 
In contrast to previous works that are trained with a single inference path, our model learns with the augmented decoding paths. 
Also, our framework can be applied to any transformer-based model. 
More explanation of HOI transformers are in Section~\ref{section3.1}.

\subsection{Consistency Learning in Vision}
\label{section2.3}
Consistency constraints applied to many computer vision topics have been extensively studied.
In semi-supervised learning, consistency regularization is widely used to train the model to be invariant to input noise.
Label consistency methods~\cite{laine2016temporal,xie2020unsupervised,miyato2018virtual,tarvainen2017mean} augment or perturb an input image and apply consistency loss between model predictions.  
CDS~\cite{NEURIPS2019_d0f4dae8} explored object detection in a semi-supervised setting with classification and localization consistency regularization. 
Also, consistency regularization in cyclic form is commonly used in generative models~\cite{zhu2017unpaired}, image matching~\cite{7298723,zhou2016learning}, temporal correspondence~\cite{dwibedi2019temporal}, and in many other domains. 


\paragraph{Comparison with Consistency Learning}
Our consistency training scheme is relevant to cross-task consistency learning~\cite{zamir2020robust}.
Cross-task consistency learning is based on inference-path invariance, where the predictions should be consistent regardless of the inference paths.


As shown in Figure~\ref{fig:consistency variants} (b), cross-\textit{task} consistency learning uses an auxiliary task $\mathcal{Y}\rightarrow \mathcal{Y}_a$ to train the main task function $f_{\mathcal{X}\mathcal{Y}}$, 
\ie, given $x$ from the query domain, and $y$ from target domain $\mathcal{Y}$, predictions of $f_{\mathcal{Y}\mathcal{Y}_a}\circ f_{\mathcal{X}\mathcal{Y}}(x)$ and $f_{\mathcal{Y}\mathcal{Y}_a}(y)$ are expected to be consistent.
Different from cross-\textit{task} consistency, our cross-\textit{path} consistency learning (Figure~\ref{fig:consistency variants} (c)) trains the main task function $f_{\mathcal{X}\mathcal{Y}}$ by enforcing the prediction of target domain $\mathcal{Y}$ of $f_{\mathcal{X}\mathcal{Y}}$ and $f_{\mathcal{Y}_s\mathcal{Y}}\circ f_{\mathcal{X}\mathcal{Y}_s}$, where auxiliary domain $\mathcal{Y}_s$ is decomposed from the target domain $\mathcal{Y}$,  to be consistent.
Also, while cross-\textit{task} consistency learning requires the mapping function $f_{\mathcal{Y}\mathcal{Y}_a}$ to be pretrained to avoid suboptimal training with the noisy estimator, cross-\textit{path} consistency learning does not demand any task-specific pre-trained networks since the auxiliary domain $\mathcal{Y}_s$ is part of the target domain $\mathcal{Y}$. 
Details for our framework is described in section~\ref{section3.2}.

\section{Method}
In this section, we present our novel end-to-end training strategy for Transformers with \textbf{cross-path consistency} in Human-Object Interaction Detection.
The training strategy includes 1) augmenting the decoding path and 2) consistency regularization between predictions of multiple decoding paths. 
Before discussing our training strategy, we briefly summarize transformers in Human-Object Interaction detection.

\subsection{Transformer in HOI detection}
\label{section3.1}
HOI transformers are commonly extended upon DETR~\cite{carion2020end}, which is composed of a CNN backbone followed by the encoder-decoder architecture of Transformer~\cite{Vaswani17transformer}. 
\jhprev{The CNN backbone first extracts a \textit{locally} aggregated feature map $f\in\R^{H'\times W' \times D}$ from input image $x \in \mathbb R^{H \times W \times 3}$.
Then, the feature map $f$ is passed into the encoder to \textit{globally} aggregate features via the self-attention mechanism, resulting in the encoded feature map $X\in\R^{H'\times W' \times D}.$
}
At a decoding stage, a decoder takes learnable query embeddings $\jhprev{q}\in \R^{N\times D}$ and outputs $e\in\R^{N\times D}$ by interacting with encoded feature map $X$ through cross-attention. 
The outputs are converted to final HOI predictions (\ie, human, object, interaction) by read-out functions, which are generally feed-forward networks. 

Training Transformers for detection entails matching between predictions and ground truth labels since Transformers provide detections as set predictions. 
To compute losses, the Hungarian algorithm~\cite{kuhn1955hungarian} is used to associate detections with ground truth labels.
The predictions unmatched with ground truth labels are considered as no object or no interactions.  
In general, HOI transformers can be categorized into two groups based on human/object localization schemes. 
\cite{tamura2021qpic,zou2021end} directly predict the box coordinates of human and object from an HOI prediction.
But this causes problems that human or object can be redundantly predicted by multiple query embeddings and the localizations of the same object often differ across HOI triplet predictions.
To address these problems, ~\cite{kim2021hotr,chen2021reformulating} propose parallel architectures to perform interaction detection separately from object detection.


\subsection{Decoding-Path Augmentation}
\label{section3.2}      
\begin{figure}[tp]
\begin{subfigure}[t]{.49\linewidth}
\includegraphics[width=\textwidth]{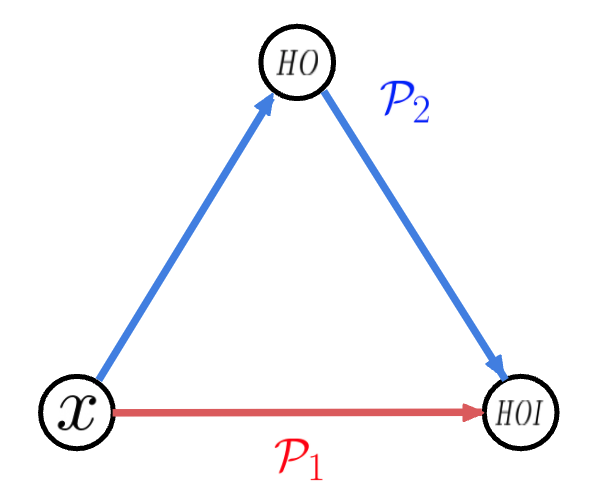}
\caption{}
\end{subfigure}
\begin{subfigure}[t]{.49\linewidth}
\includegraphics[width=\textwidth]{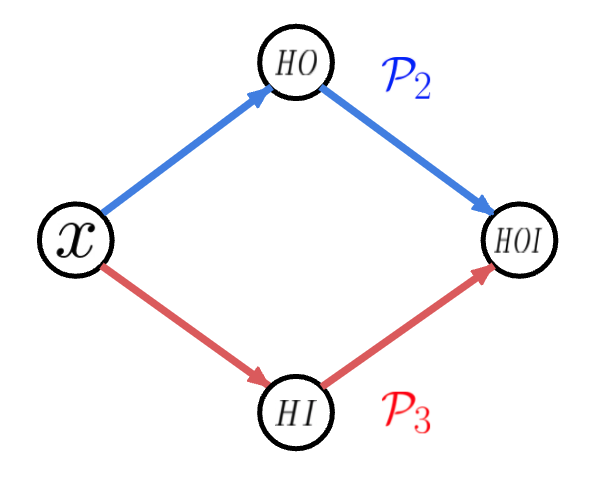}
\caption{}
\end{subfigure}
\caption{\textbf{Cross-path consistency for HOI detection}. (a) Main task path ${\color{red}\mathcal{P}_1}$ should be consistent with each augmented path. \eg path ${\color{blue}\mathcal{P}_2}$. (b) Augmented paths should be consistent with one another. \eg path ${\color{blue}\mathcal{P}_2}$ and ${\color{red}\mathcal{P}_3}$. }
\label{fig:consistency for HOI}
\end{figure}
\begin{figure*}[t]
    \centering
    \includegraphics[width=0.9\textwidth]{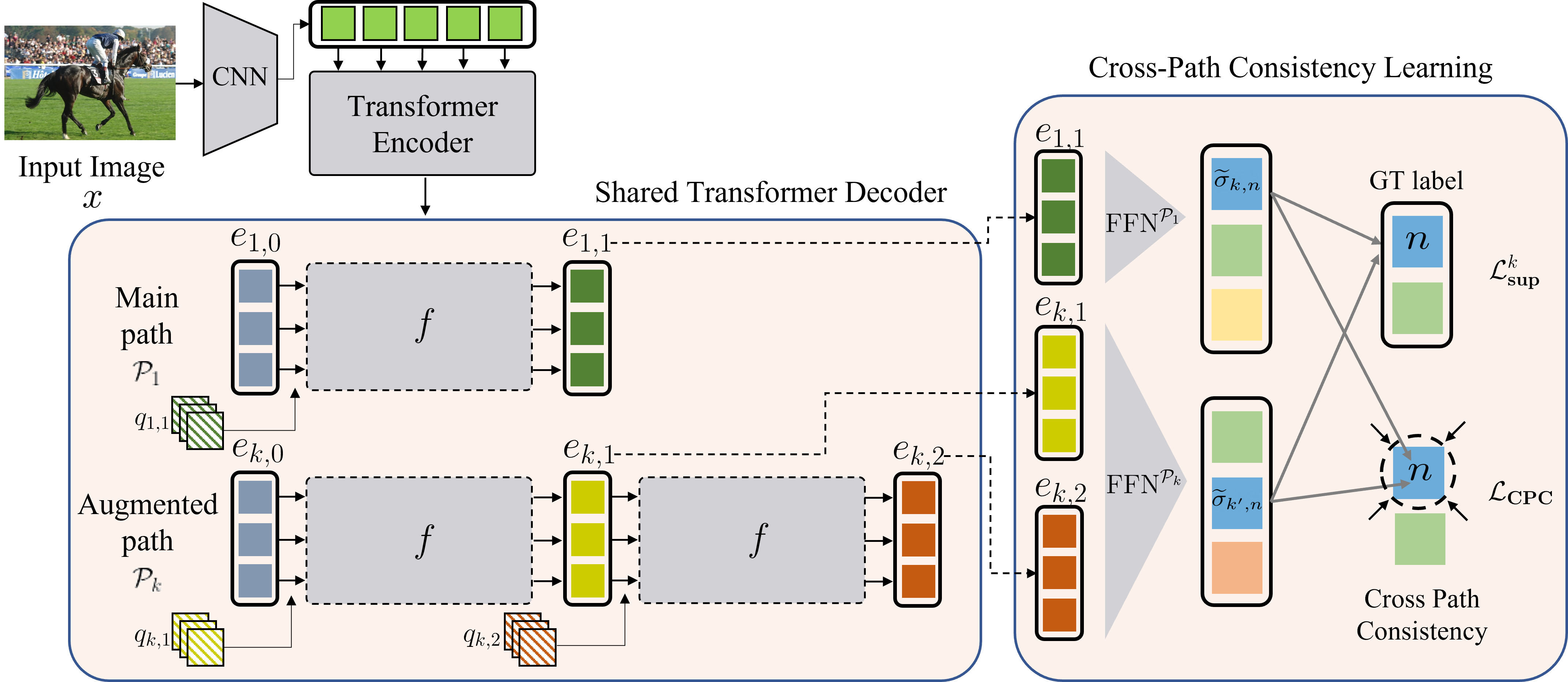}
    \caption{
    \label{fig:pipeline}
    {\textbf{The overall process of Cross-Path Consistency Learning}. The encoded image features are passed into the shared decoder with multiple inference paths $\{\paths_1, ..., \paths_{k-1}, \paths_k\}$. 
    Each path is augmented based on the decoding-path augmentation to generate various sequences of inference paths (see Section~\ref{section3.2}). 
    To avoid clutter, we visualize only the main path $\paths_1$ and an augmented path $\paths_k$. 
    The main path $\paths_1$ consists of a single decoding stage,
    and the augmented path $\paths_k$ is a composition of decoding stages; all $f$ blocks share parameters.
    Given queries $q$ a learnable position embeddings, each decoder extracts output embeddings denoted as $e_{1,1}$, $e_{k,1}$, and $e_{k,2}$. 
    Then, each of the output embeddings is fed into the readout function $\text{FFN}$ to predict each HOI element \ie \texttt{<human, object, interaction>}. 
    With Cross-Path Consistency Learning (Section ~\ref{section3.3}), all the outputs supervised with the same ground truth label are trained to be consistent regardless of their inference paths. 
    Cross-Matching is used to match the queries that are considered to be consistent by leveraging ground truth label. 
    Along with the supervision loss $\mathcal{L}_\mathbf{sup}^k$ for all paths $\paths_k$, cross-path consistency loss $\mathcal{L}_\textbf{CPC}$ is added to our final loss.}}
    \label{fig1}
\end{figure*}

We observe that HOI detection can be achieved by various sequences of predictions.
For instance, CNN-based HOI detection models~\cite{gkioxari2018detecting, chao2018learning, Gao2020DRG, gupta2019no} first detect instances (human and object) and then predict interactions between the instances, 
\ie, $x \rightarrow \text{HO} \rightarrow \text{I}$, where $x$ is an input image and H, O, I are predictions for human, object, interaction, respectively. 
On the other hand, the HOI Transformers by \cite{kim2021hotr,tamura2021qpic, chen2021reformulating, zou2021end} directly predict HOI triplets, \ie, $x \rightarrow \text{HOI}$.
Inspired by Cross-Task Consistency \cite{zhang2018deep} and this observation, 
we propose \textbf{decoding-path augmentation} to generate various decoding paths (or prediction paths) and impose consistency regularization.
Decoding-path augmentation for Transformers in HOI detection can be easily achieved by partially decoded HOI predictions.
Furthermore, sharing decoders across paths is beneficial in terms of knowledge sharing.

In our experiments, we consider four decoding paths as follows:
\begin{equation}
\begin{split}
    \label{eq:all_path}
    &\;\paths_1 = x \rightarrow \text{HOI} \\ 
    &\begin{rcases*}
    \paths_2 =x \rightarrow \text{HO} \rightarrow \text{I} \\ 
    \paths_3 = x\rightarrow \text{HI} \ \rightarrow \text{O} \\ 
    \paths_4 =x \rightarrow \text{OI} \ \rightarrow \text{H}
    \end{rcases*}
    Augmented.
\end{split}
\end{equation}


Each decoding stage of path $\paths_k$ can be written as:  
\begin{equation}
    \label{eq:path_decoding}
    \begin{split}
        e_{{k,1}} &= f(e_{k,0}+ q_{k,1}, \ X), \\
        e_{{k,2}} &= f(e_{k,1} + q_{k,2}, \ X),
    \end{split}
\end{equation}
where ${q}_{k,j}, \ e_{k,j}$ denote learnable query and output embeddings on $k^{\text{th}}$ path at $j^{\text{th}}$ decoding stage. The decoder $f$ is shared across all paths and stages.
The $e_{k,0}$ above is dummy output embeddings set to zeros since there is no 0-th stage, see Figure \ref{fig:pipeline}.
Each decoding stage and path use a separate readout function $\text{FFN}$ to translate the output embeddings into HOI instance predictions. 
For example, on $\paths_2: x \rightarrow \text{HO} \rightarrow \text{I}$, at stage 1
$e_{2,1}$ is read out by $\text{FFN}^{\paths_2}_{h}$ and $\text{FFN}_{o}^{\paths_2}$ to predict bounding boxes of human and object respectively.
 Prediction for HOI element \jhprev{$m \in \{h,o,act \}$} in each $k^{th}$ path at $j^{th}$ decoding stage can be written as $\hat{y}_{k}^m=\text{FFN}^{\paths_k}_m\left(e_{k,j}\right)$.

\subsection{Cross-Path Consistency Learning}
\label{section3.3}
\hjk{We now present our Cross-Path Consistency Learning framework (CPC) that imposes consistency regularization between predictions from different decoding paths as shown in Figure~\ref{fig:consistency for HOI}.
Learning with CPC leads better generalization without any additional data or labels. 

\paragraph{Cross-Path Consistency.}
We explain our consistency learning scheme with an exemplary case of main path $\paths_1$ and augmented path $\paths_2$ given as
\begin{equation}
\label{eq:path}
    \begin{split}
        &\paths_1 : x \rightarrow \text{HOI} \\ 
        &\paths_2 : x \rightarrow \text{HO} \rightarrow \text{I}. 
    \end{split}
\end{equation} 
Here, the main path $\paths_1$ is the HOI transformers' original inference path.
In path $\paths_2$, human and object detection logits $\hat{y}^h_2$ and $\hat{y}^o_2$ are obtained reading out $e_{2,1}$, which is the output embeddings on path 2 at stage 1. 
Then, the interaction logit $\hat{y}^{act}_2$ is obtained after another subsequent decoder pass defined as $f_{2,2}$.
The corresponding inference scheme of $\paths_2$ can be written in more formal terms:
\begin{equation}
\begin{split}
\label{eq:out2}
    \hat{y}_2^{h}&=\text{FFN}_h^{\paths_2}(f_{2,1}(X)) \\
    \hat{y}_2^{o}&=\text{FFN}_o^{\paths_2}(f_{2,1}(X)) \\
    \hat{y}_2^{act}&=\text{FFN}_{act}^{\paths_2}(f_{2,2}\circ f_{2,1}(X)) \\
\end{split}
\end{equation}
In ~\eqref{eq:out2}, input arrays for $f$ other than feature map $X$ were omitted for simplicity.

With the predictions, we impose regularization to make the outputs from path $\paths_1$ and path $\paths_2$ consistent. 
Note that HOI detections from $\paths_2$ consist of both final and intermediate decoder outputs.
To this end, we define the loss function $\mathcal{L}_{\paths_{1}\paths_2}$ by aggregating losses from multiple augmented paths to enforce consistency. 
The loss function is given as:
\begin{equation}
\begin{split}
     \mathcal{L}_{\paths_{1}\paths_2}=\lambda_h\cdot\mathcal{L}_{h}\big(\hat{y}_1^{h},\hat{y}_2^{h}\big)  +\lambda_o\cdot
     \mathcal{L}_{o}\big(\hat{y}_1^{o},\hat{y}_2^{o}\big) \\ +\lambda_{act}\cdot\mathcal{L}_{act}\big(\hat{y}_1^{act},\hat{y}_2^{act}\big),
\end{split}
\end{equation}
where $\hat{y}^h_1$, $\hat{y}^o_1$ and $\hat{y}^{act}_1$ are the output from the main path $\paths_1$ and $\lambda$ are the loss weights. 
In our experiments, softmax-type outputs use Jensen-Shannon divergence (JSD) for consistency loss to give loss to each path symmetrically, while outputs followed by sigmoid, \eg, box regression, multi-label action classes, take the Mean-Squared Error loss. 
More details on type-specific loss functions are in the supplement.

In the case of other path pairs, loss is computed in the same manner. The final loss should thus incorporate all possible pairs. Then, the cross-path consistency (CPC) loss can be written as:
\begin{equation}
\label{eq:cpcloss}
    \mathcal{L}_\textbf{CPC}=\frac{1}{S}\sum_{(k,k')\in
\mathcal{K}}\mathcal{L}_{\paths_k\paths_{k'}}
\end{equation}
where $\mathcal{K}$ denotes the set of all possible path pairs, and $S$ refers to the size of set $\mathcal{K}$, \ie the number of path combinations.}

\paragraph{Cross Matching.}
\hjk{
Cross-path consistency learning encourages outputs from different paths to be consistent.
However, since the outputs from a path are given as a set, we first need to 
resolve correspondence to specify the pairs of predictions to enforce consistency.
We present \textbf{cross matching}, a simple method that tags each instance with its corresponding ground truth label. 
The instances tagged with the same label are paired to compute consistency loss.
On the other hand, if an instance is not matched with any of the paths' output, 
we simply exclude the instance from consistency learning treating it as \textit{no object} or \textit{no interaction}.
Our cross-path consistency loss is introduced below.
}



Let $\sigma_k(i)$ denote the index of the ground truth label that matches the $i^{th}$ query in the $k^{th}$ path. 
We define $\sigma_{k}^{-1}\left(n\right)$ as the query index of path $\paths_k$ which is matched with the ground truth index $n$.
To avoid clutter, we use  $\tilde\sigma_{k,n}$ as a shorthand notation for $\sigma_{k}^{-1}\left(n\right)$.
The outputs from different paths with the same ground-truth label should be consistent. 
For example, $\hat y^m_{k, {\tilde{\sigma}_{k,n}}}$ and $\hat y^m_{k, {\tilde{\sigma}_{k^\prime,n}}}$ which are predictions for $m$ from $\paths_k$ and $\paths_{k^\prime}$ with the same ground-truth index $n$ should be consistent.   

Cross-path consistency loss between output predictions from $\paths_k$ and $\paths_{k'}$ with the same ground-truth with index $n$ is defined as follows:
\begin{equation}
\begin{split}
    \mathcal{L}_{\paths_{k}\paths_{k'}}^n
    =  &\hspace{2mm} \lambda_h\cdot\mathcal{L}_{h}\big(\hat{y}^{h}_{k,\tilde\sigma_{k,n}},\hat{y}^{h}_{k',\tilde\sigma_{k',n}}\big) \\
    & + \lambda_o\cdot\mathcal{L}_{o}\big(\hat{y}^{o}_{k,\tilde\sigma_{k,n}},\hat{y}^{o}_{k',\tilde\sigma_{k',n}}\big) \\ 
    & + \lambda_{act}\cdot \mathcal{L}_{act}\big(\hat{y}^{act}_{k,\tilde\sigma_{k,n}},\hat{y}^{act}_{k',\tilde\sigma_{k',n}}\big)  
\end{split}
\end{equation}

\paragraph{Final Loss}
The final cross-path consistency loss for all $\paths_k$ is derived as,
\begin{equation}
    \mathcal{L}_{\textbf{CPC}}=\frac{1}{{S}\cdot\mathcal{N}}\sum_{n=1}^\mathcal{N}\sum_{(k,k')\in \mathcal{K}}\mathcal{L}_{\paths_{k}\paths_{k'}}^n
\end{equation}
where $\mathcal{N}$ is the number of ground truth labels.
Then, the final form of our training loss $\mathcal{L}$ is defined by
\begin{equation}
    \mathcal{L}=\sum_k\mathcal{L}_{\mathbf{sup}}^k+w(t)\cdot\mathcal{L}_{\textbf{CPC}},
\end{equation}
where $\mathcal{L}_\mathbf{sup}^k$ is the supervision loss for each path $\paths_k$ and $w(t)$ is a ramp-up function~\cite{laine2016temporal,berthelot2019mixmatch,tarvainen2017mean} for stable training.
Our overall framework is illustrated in Figure ~\ref{fig1}.


\section{Experiments}

In this section, we empirically evaluate the effectiveness of our cross-path consistency learning with HOI transformers. 
Our experiments are conducted on public HOI detection benchmark datasets: \textbf{V-COCO} and \textbf{HICO-DET}.
We first briefly introduce the datasets and provide implementation details.
Our extensive experiments demonstrate that our training strategy renders significant improvement on the baseline models without additional parameters or inference time. 

\subsection{Dataset}
\paragraph{V-COCO}~\cite{gupta2015visual} is a subset of the COCO dataset~\cite{lin2014mscoco} which contains 5,400 \texttt{trainval} images and 4,946 \texttt{test} images.
V-COCO is annotated with 29 common action classes.
For evaluation of the V-COCO dataset, we report the mAP metric over 25 interactions for two scenarios, 
The first scenario includes a prediction of occluded objects and is evaluated with respect to {AP}${}^{}_{\text{role1}}$. 
On the other hand, the second scenario does not contain such cases, and performance is measured in {AP}${}^{}_{\text{role2}}$.

\paragraph{HICO-DET}~\cite{chao2018learning} is a subset of the HICO dataset which has more than 150K annotated instances of human-object pairs in 47,051 images (37,536 for training and 9,515 for testing). 
It is annotated with 600 $\texttt{<interaction, object>}$ instances.
There are 80 unique object types, identical to the COCO object categories, and 117 unique interaction verbs.
For evaluation of the HICO-DET, we report the mAP over three different set categories: (1) all 600 HOI categories in HICO (Full), (2) 138 HOI categories with less than 10 training samples (Rare), and (3) 462 HOI categories with more than 10 training samples (Non-Rare).
\subsection{Implementation Details}
\paragraph{Training}
In our experiment, QPIC~\cite{tamura2021qpic} and HOTR~\cite{kim2021hotr} were used as the baseline for the HOI transformer respectively.
During training, we initialize the network with pretrained DETR~\cite{carion2020end} on MS-COCO with a Resnet-50 backbone.
For all decoding paths, parameters of the model are shared except for stage-wise queries and feedforward networks.

All our experiments using consistency regularization are trained for 90 epochs and the learning rate is decayed at the 60-th epoch by a factor of 0.1. 
As an exception, HOTR is trained up to 50 epochs and the learning rate is decayed at epoch 30 by a factor of 0.1  for HICO-DET.
Following the original training schemes in QPIC and HOTR, we freeze the encoder and backbone for HOTR, whereas unfreeze those for QPIC.
We use the AdamW~\cite{loshchilov2017decoupled} optimizer with a batch size of 16, and the initial learning rates for the transformer and backbone parameters are set to $10^{-4}$ and $10^{-5}$ respectively, and weight decay is set to $10^{-4}$. 
All experiments are trained on 8 V100 GPUs.

We re-implement the result of QPIC and HOTR on V-COCO~\cite{gupta2015visual}  since our reproduction results are quite different from the official ones in the paper.
For a fair comparison, all the loss coefficients overlapping between baselines and our training strategy are identical to the ones reported in the paper~\cite{kim2021hotr,tamura2021qpic}.
Details for hyperparameters relevant to our training strategy are reported in the supplementary material.
\paragraph{Inference} %
We mainly use $\paths_1$ ( $x\rightarrow \text{HOI}$ ) for inference to compare with the baseline models without increasing the number of parameters.
Also, we report the results of other inference paths in our ablation studies. 


\subsection{Comparison with HOI transformer}

    
         
         
         
         

    
We evaluate the effectiveness of our method compared to the existing HOI transformers.
All experiments are reported with the main path $\paths_1$ that infers HOI triplets by a single decoding stage ($x\rightarrow\text{HOI}$) which is identical to the original HOI transformer.
As shown in Table~\ref{table:comp}, our CPC training strategy significantly outperforms on two baselines, HOTR~\cite{kim2021hotr} and QPIC~\cite{tamura2021qpic}.
In the V-COCO dataset, the experiment shows improvement in performance by a considerable margin of 0.9 mAP for QPIC in {AP}${}^{}_{\text{role1}}$, and 1.8 mAP for HOTR.
For {AP}${}^{}_{\text{role2}}$, QPIC and HOTR gain improvement by 0.9 mAP and 1.9 mAP respectively, similar to that of AP$_\text{role1}$.  

In the HICO-DET dataset, our CPC learning with HOTR and QPIC outperforms all the evaluation categories of HICO-DET, except negligible degradation in the Non-Rare category on HOTR.
Results on rare class on the HICO-DET are improved by a significant margin of 1.29 mAP and 5.5 mAP for QPIC and HOTR respectively.
In both models, we observe a more prominent performance improvement in the Rare category.
This supports that our training strategy performs well on rarely seen examples.
Our strategy improves the conventional HOI transformer models.

\begin{table}[!ht]
    \centering
    \footnotesize
    
    \begin{tabular}{l  c c  c  c c }
         \toprule
         \multicolumn{1}{c}{} &\multicolumn{2}{c}{V-COCO}& \multicolumn{3}{c}{HICO-DET} \\
         \cmidrule(r){2-3}
         \cmidrule(r){4-6}
         Method & 
         \scone & 
         \sctwo & 
         $\  $ {Full} $\  $ & 
         {Rare} & 
         {Non-Rare   }
         \\
         \midrule
         \midrule
         
        
        QPIC & 62.2* & 64.5* &29.07 &21.85& 31.23 \\
        QPIC + ours & \textbf{63.1} & \textbf{65.4}& \textbf{29.63}&\textbf{23.14} &\textbf{31.57}  \\
         \midrule
         
         
         HOTR& 59.8* & 64.9* & 25.10& 17.34&\textbf{27.42}  \\
         HOTR + ours& \textbf{61.6}&\textbf{66.8} &\textbf{26.16} &\textbf{22.84} &27.15  \\
         \bottomrule
    \end{tabular}
    \caption{\textbf{Comparison of our training strategy with vanilla HOI transformers on V-COCO and HICO-DET}. * signifies our results reproduced with the official implementation codes of QPIC and HOTR.}
    \label{table:comp}
    
\end{table}

\begin{table}[ht]
  \centering
  \footnotesize
  \begin{tabular}{l|c|c c}
    \toprule
    Method\hspace{15pt} & \hspace{15pt}Backbone\hspace{15pt} & \scone & 
    \sctwo  \\ \midrule \midrule
     \multicolumn{4}{l}{ {\textbf{CNN-based HOI Detection Model}}}\vspace{0.3pt} \\ \midrule
    InteractNet~\cite{gkioxari2018detecting} & R50-FPN & 40.0 & 48.0 \\
    iCAN~\cite{gao2018ican} & R50 & 45.3 & 52.4 \\
    TIN~\cite{li2019transferable} & R50  & 47.8 & - \\
    RPNN~\cite{zhou2019relation} & R50  & - & 47.5 \\
    Verb Embd.~\cite{xu2019learning} & R50 & 45.9 & - \\
    PMFNet~\cite{wan2019pose} & R50-FPN  & 52.0 & - \\
    PastaNet~\cite{li2020pastanet} & R50-FPN  & 51.0 & 57.5 \\ 
    VCL~\cite{Hou2020VCL}  & R50 L & 48.3 & - \\
    UniDet~\cite{Kim2020UnionDet} & R50-FPN  & 47.5 & 56.2 \\ 
    DRG~\cite{Gao2020DRG} & R50-FPN  & 51.4 & - \\
    FCMNet~\cite{Liu2020FCMNet} & R50  & 53.1 & - \\
    ConsNet~\cite{Liu2020ConsNet} & R50-FPN  & 53.2 & - \\
    PDNet~\cite{Xubin2020PDNet} & R50-FPN  & 53.3 & - \\
    IDN~\cite{yonglu2020idn} & R50  & 53.3 & 60.3 \\ 
    GPNN~\cite{qi2018learning} & R152 & 44.0 & - \\
    IPNet~\cite{wang2020learning} & H.G.104 & 51.0 & - \\
    VSGNet~\cite{ulutan2020vsgnet} & R152  & 51.8 & 57.0 \\ 
    PDNet~\cite{Xubin2020PDNet}  & Res152  & 52.2 & - \\ 
    ACP~\cite{Kim2020ACP} & Res152  & 53.0 & - \\  \midrule
     \multicolumn{4}{l}{ {\textbf{Transformer-based HOI Detection Model}}}  \\  \midrule 
    HoiT ~\cite{zou2021end} & R101  & 52.9 & - \\
    AS-Net ~\cite{chen2021reformulating} & R50  & 53.9 & - \\
   
    \hline
    HOTR ~\cite{kim2021hotr} & R50  & 55.2 & 64.4 \\
    HOTR+ \textbf{Ours} & R50 & \textbf{61.6} & \textbf{66.8}  \\
    QPIC ~\cite{tamura2021qpic} & R50  & 58.8 & 61.0 \\
    QPIC+ \textbf{Ours} & R50 & \textbf{63.1} &\textbf{ 65.4} \\
    \bottomrule
  \end{tabular}
  \vspace{5pt}
  \caption{\textbf{Comparison of performances on the V-COCO test set}. {AP}${}^{}_{\text{role1}}$  and {AP}${}^{}_{\text{role2}}$ denotes performances under Scenario 1 and Scenario 2 in V-COCO respectively.} 
  \label{tab:V-COCO}
\end{table}

\begin{table*}[!ht]
  \centering
  \footnotesize
  \begin{tabular}{l c c c c c c}
    \toprule
    \multicolumn{4}{c}{} & \multicolumn{3}{c}{\textbf{Default}} \\
    \cmidrule(r){5-7}
    Method & \hspace{12pt}Detector\hspace{12pt} & Backbone & \hspace{12pt}Extra\hspace{12pt} & \hspace{12pt}Full\hspace{12pt} & \hspace{6pt}Rare\hspace{6pt} & Non Rare\hspace{6pt} \\ \midrule\midrule
     \multicolumn{3}{l}{ {\textbf{CNN-based HOI Detection Model}}} &  &  &  &  \\ \midrule
    \multicolumn{1}{l|}{InteractNet~\cite{gkioxari2018detecting}} & COCO & R50-FPN & \multicolumn{1}{c|}{\xmark} & 9.94 & 7.16 & 10.77  \\
    \multicolumn{1}{l|}{iCAN~\cite{gao2018ican}} & COCO & R50 & \multicolumn{1}{c|}{S} & 14.84 & 10.45 & 16.15  \\
    \multicolumn{1}{l|}{TIN~\cite{li2019transferable}} & COCO & R50 & \multicolumn{1}{c|}{S+P} & 17.03 & 13.42 & 18.11  \\
    \multicolumn{1}{l|}{RPNN~\cite{zhou2019relation}} & COCO & R50 & \multicolumn{1}{c|}{P} & 17.35 & 12.78 & 18.71  \\
    \multicolumn{1}{l|}{PMFNet~\cite{wan2019pose}} & COCO & R50-FPN & \multicolumn{1}{c|}{S+P} & 17.46 & 15.65 & 18.00  \\
    \multicolumn{1}{l|}{No-Frills HOI~\cite{gupta2019no}} & COCO & R152 & \multicolumn{1}{c|}{S+P} & 17.18 & 12.17 & 18.68  \\
    \multicolumn{1}{l|}{UnionDet~\cite{Kim2020UnionDet}} & COCO & R50-FPN & \multicolumn{1}{c|}{\xmark} & 14.25 & 10.23 & 15.46 \\
    \multicolumn{1}{l|}{DRG~\cite{Gao2020DRG}} & COCO & R50-FPN & \multicolumn{1}{c|}{S+L} & 19.26 & 17.74 & 19.71  \\
    \multicolumn{1}{l|}{VCL~\cite{Hou2020VCL}} & COCO & R50 & \multicolumn{1}{c|}{S} & 19.43 & 16.55 & 20.29  \\
    \multicolumn{1}{l|}{FCMNet~\cite{Liu2020FCMNet}} & COCO & R50 & \multicolumn{1}{c|}{S+P} & 20.41	& 17.34	& 21.56  \\
    \multicolumn{1}{l|}{ACP~\cite{Kim2020ACP}} & COCO & R152 & \multicolumn{1}{c|}{S+P} & 20.59 & 15.92 & 21.98  \\
    \multicolumn{1}{l|}{DJ-RN~\cite{li2020detailed}} & COCO & R50 & \multicolumn{1}{c|}{S+V} & 21.34 & 18.53 & 22.18  \\
    \multicolumn{1}{l|}{ConsNet~\cite{Liu2020ConsNet}} & COCO & R50-FPN & \multicolumn{1}{c|}{S+L} & 22.15 & 17.12 & 23.65  \\
    \multicolumn{1}{l|}{PastaNet~\cite{li2020pastanet}} & COCO & R50 & \multicolumn{1}{c|}{S+P+L} & 22.65 & 21.17 & 23.09  \\
    \multicolumn{1}{l|}{IDN~\cite{yonglu2020idn}} & COCO & R50 & \multicolumn{1}{c|}{S} & 23.36 & 22.47 & 23.63  \\ 
    
    \multicolumn{1}{l|}{GPNN~\cite{qi2018learning}} & COCO & R152 & \multicolumn{1}{c|}{\xmark} & 13.11 & 9.41 & 14.23  \\
    \multicolumn{1}{l|}{IPNet~\cite{wang2020learning}} & COCO & HourGlass104 &\multicolumn{1}{c|}{\xmark} & 19.56 & 12.79 & 21.58 \\
    \multicolumn{1}{l|}{VSGNet~\cite{ulutan2020vsgnet}} & COCO & R152 & \multicolumn{1}{c|}{S} & 19.80 & 16.05 & 20.91  \\
    \multicolumn{1}{l|}{PD-Net~\cite{Xubin2020PDNet}} & COCO & R152 & \multicolumn{1}{c|}{S+P+L} & 20.81 & 15.90 & 22.28  \\ \midrule 
    
     \multicolumn{3}{l}{{ \textbf{Transformer-based HOI Detection Model}}} &  &  &  &   \\ \midrule
    \multicolumn{1}{l|}{HoiT~\cite{zou2021end}} & HICO-DET & R50 & \multicolumn{1}{c|}{\xmark} & 23.46 & 16.91 & 25.41 \\
    \multicolumn{1}{l|}{AS-Net~\cite{chen2021reformulating}} & HICO-DET & R50 & \multicolumn{1}{c|}{\xmark} & 28.87 & 24.25 & 30.25 \\
     \hline 
    \multicolumn{1}{l|}{HOTR~\cite{kim2021hotr}} & HICO-DET & R50 & \multicolumn{1}{c|}{\xmark} & 25.10 & 17.34 & \textbf{27.42} \\
    \multicolumn{1}{l|}{HOTR+ \textbf{Ours}} & HICO-DET & R50 & \multicolumn{1}{c|}{\xmark} & \textbf{26.16} & \textbf{22.84} & 27.15 \\
    \multicolumn{1}{l|}{QPIC~\cite{tamura2021qpic}} & HICO-DET & R50 & \multicolumn{1}{c|}{\xmark} & 29.07 & 21.85 & 31.23 \\
    \multicolumn{1}{l|}{QPIC+ \textbf{Ours}} & HICO-DET & R50 & \multicolumn{1}{c|}{\xmark} & \textbf{29.63} & \textbf{23.14} & \textbf{31.57} \\
    \bottomrule
  \end{tabular}
  \vspace{5pt}
  \caption{\textbf{Performance comparison in HICO-DET}. For the Detector, COCO means that the detector is trained on COCO, while HICO-DET means that the detector is first trained on COCO and then fine-tuned on HICO-DET.  The each letter in Extra column stands for S: Interaction Patterns (Spatial Correlations), P: Pose, L: Linguistic Priors, V: Volume.}
  \label{tab:HICO-DET}
\end{table*}
\begin{table*}[ht]
    \centering
    \small
    \begin{tabular}{c c c   c c c c c}
         \toprule
         \multirow{1}{*}{\textbf{Method}} & 
         \multirow{1}{*}{\textit{Share Dec.}} &
         \multirow{1}{*}{$\  $\textit{CPC}$\  $}& \multicolumn{1}{c}{$\paths_1$}
         &\multicolumn{1}{c}{$\paths_2$}
         &\multicolumn{1}{c}{$\paths_3$}
         &\multicolumn{1}{c}{$\paths_4$}
         &\multicolumn{1}{c}{\textbf{\textit{Average}}} \\
         
         \midrule    
         \midrule
         
         \multirow{3}{*}{QPIC} 
         
         & \checkmark
         &\checkmark & 
         \textbf{63.1} & 
         \textbf{63.3} &  
         \textbf{63.1} & 
         \textbf{63.0} & 
         \textbf{63.13} $\pm$ 0.05${}^\dagger$ \\
         
         & 
         &\checkmark & 
         62.4 & 
         62.9 &  
         60.8 & 
         59.4 & 
         61.38 $\pm$ 1.38 \\
         
         & \checkmark
         &  &
         60.7 &
         60.7 &
         59.9 & 
         58.1 &
         59.85 $\pm$ 1.06 
         \\
         
         \midrule

         \multirow{3}{*}{HOTR} 
         
         & \checkmark
         &\checkmark &
         \textbf{61.6} &  
         61.5 & 
         \textbf{61.6} &  
         \textbf{61.6} &  
         \textbf{61.58} $\pm$ 0.02${}^\dagger$  
         \\
         
         & 
         &\checkmark &
         61.2 &  
         \textbf{61.6} & 
         61.1 &  
         60.6 &  
         61.13 $\pm$ 0.36  
         \\
         
         & \checkmark& & 
         60.6 & 
         60.6 &  
         61.2 &  
         60.6 &  
         60.75 $\pm$ 0.13 
         \\

         \bottomrule
    \end{tabular}
    \caption{
    \textbf{Ablation Study on our learning strategies}. Ablation results on shared decoder (\textit{Share Dec.}), and Cross-Path Consistency (CPC) are demonstrated. For main path $\mathcal{P}_1$, and each augmented path $\mathcal{P}_2$, $\mathcal{P}_3$, $\mathcal{P}_4$, their performances are reported measured in mAP. They are evaluated on the V-COCO test set with respect to Scenario 1. The best performances for each path are highlighted in bold, and $\dagger$ refers to the case where the least standard deviation is observed. 
    }
    \label{table:abl}
\end{table*}

\subsection{Comparison with State-of-the-Art Methods}
In Table~\ref{tab:V-COCO} and Table~\ref{tab:HICO-DET}, we compare previous HOI detection methods with ours.
As demonstrated in the tables, our training strategy achieves the best performance among its peers.
Table~\ref{tab:V-COCO} shows the result on V-COCO dataset in both {AP}${}^{}_{\text{role1}}$ and {AP}${}^{}_{\text{role2}}$.
In the V-COCO dataset, our method achieves outstanding performance of 63.1 mAP in {AP}${}^{}_{\text{role1}}$ and 66.8 mAP in {AP}${}^{}_{\text{role2}}$. 
Also, the results on the HICO-DET dataset in Table~\ref{tab:HICO-DET} show that our CPC further improves the state-of-the-art models (\eg, HOTR, and QPIC) in the default setting achieving 26.16 mAP and 29.63 mAP, respectively.

\subsection{Ablation Study}
We further discuss the effectiveness of our framework through a series of ablation studies.
We first provide a path-wise analysis for our cross-path consistency learning method.
The effect of our training technique components was tested on each path to validate our method.
Subsequently, we analyze the impact of the number of augmented paths on the main task performance. 
We experimentally prove the validity of our method by demonstrating the correlation between the number of paths and performance.

\paragraph{Efficiency of CPC.}
Table~\ref{table:abl} presents ablation experiment results for all inference paths, $\paths_1$, $\paths_2$, $\paths_3$, and $\paths_4$.
Path $\paths_1$ is the main path, which we aim to boost performance with the rest of the augmented paths.
We try ablating decoder sharing or cross-path consistency regularization one at a time to confirm each component's contribution to our training strategy.
Note that all of our experiments are conducted with the encoder block shared across paths.

When our CPC training strategies are applied, QPIC and HOTR achieve an mAP of 63.1, and 61.6 on main path $\paths_1$.
When the decoder parameters are not shared, performance degradation in path $\paths_1$ was observed for both baselines; a 0.7 mAP drop for QPIC, and 0.4  mAP drop for HOTR. 
On the other hand, when CPC regularization is left out while decoder parameters are shared, performance of QPIC and HOTR decreased by a large margin of 2.4 mAP and 1.0 mAP each. 
In terms of overall performance across all paths, the average mAP showed a similar trend for each experiment condition.
The overall results support that our learning strategy improves generalization of base architectures, and boosts performance by sharing knowledge throughout paths and stages.

Interestingly, the standard deviation of all performances dramatically increases without both components.
With unshared decoders, deviation increases by 1.33 for QPIC and 0.35 for HOTR.
Also, when CPC regularization is removed, deviation increases by 1.01 for QPIC and 0.11 for HOTR. 
This implies that our training strategy with shared decoder and CPC leads to more stable training as well as consistent representations.

\paragraph{Impact of Augmented Paths.}
We explore how the number of augmented paths affects the performance of the main path $\paths_1$ in V-COCO benchmark.
Starting from $\paths_1$, the augmented paths are gradually added with respect to mAP of Scenario 1 from Table~\ref{table:abl_pathwise_sep}, where each path is independently trained with default settings with no training techniques applied.
We leverage the augmented path with better performance first, as performance of each model will serve as a lower bound for the ensemble of paths.
Specifically, as shown in Table~\ref{table:abl_pathwise_sep}, both HOTR and QPIC showed better performance in the order of $\paths_1$, $\paths_2$, $\paths_3$, and $\paths_4$, when trained independently.

\begin{figure}[h]
\centering
\begin{subfigure}[t]{0.49\linewidth}
\includegraphics[width=\textwidth]{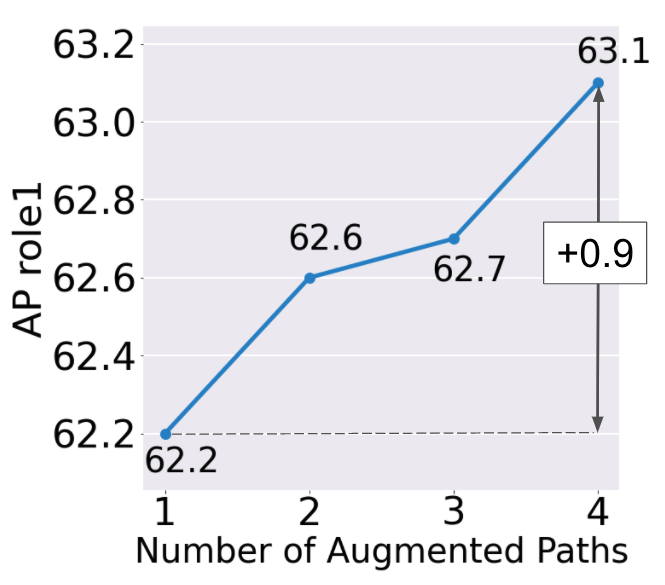}
\caption{QPIC}
\end{subfigure}
\begin{subfigure}[t]{0.49\linewidth}
\includegraphics[width=\textwidth]{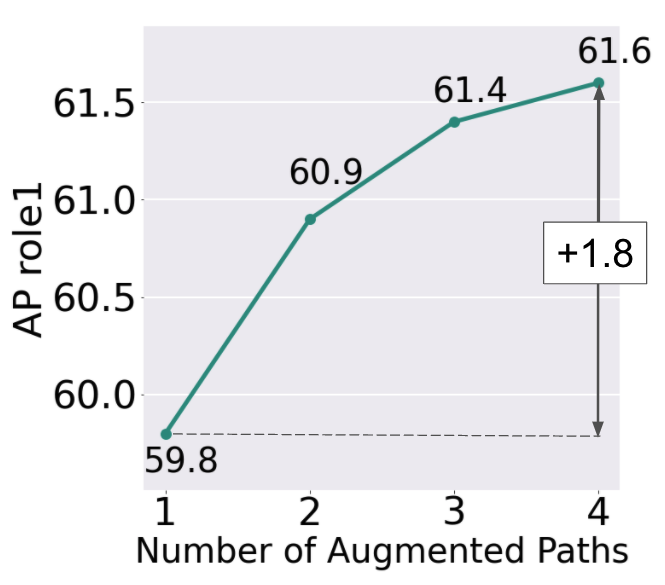}
\caption{HOTR}
\end{subfigure}
\caption{\textbf{Ablation on the number of augmented paths}. As the number of augmented paths increases, main task performance increases accordingly.}
\vspace{-14pt}
\label{fig:pathnum ablation}
\end{figure}

We compare the four cases where the augmented paths are gradually added in the corresponding order; \ie, $\paths_1$, $\paths_1+\paths_2$, $\paths_1+\paths_2+\paths_3$, and $\paths_1+\paths_2+\paths_3+\paths_4$.
As shown in Figure~\ref{fig:pathnum ablation}, performance is gradually improved as augmented paths are  added.
The ablation study evidences that regardless of each path performance, taking advantage of more paths bolsters the learning capability of our main task, and its performance builds up as the number of augmented paths increases.
\begin{table}[h]
    \centering
    \small
    \begin{tabular}{c c c c c c}
         \toprule
         \multirow{1}{*}{\textbf{Method}} &  \multicolumn{1}{c}{$\paths_1$}
         &\multicolumn{1}{c}{$\paths_2$}
         &\multicolumn{1}{c}{$\paths_3$}
         &\multicolumn{1}{c}{$\paths_4$}
         &\multicolumn{1}{c}{\textbf{\textit{Average}}} \\
         
         \midrule    
         \midrule
         
         \multirow{1}{*}{QPIC} &
         62.2 & 
         61.9 &
         61.7 &
         60.4 &
         61.55 $\pm$ 0.69 
         \\
         
         \midrule
        
         \multirow{1}{*}{HOTR}&
         59.8 &  
         59.5 & 
         59.0 & 
         58.9 & 
         59.3 $\pm$ 0.37  
         \\

         \bottomrule
    \end{tabular}
    \caption{
    \textbf{Path-wise results on V-COCO.}
    }
    \label{table:abl_pathwise_sep}
\end{table}
\vspace{-1.5mm}
\section{Conclusion}
 
We propose end-to-end Cross-Path Consistency learning for Human-Object Interaction detection. Through decoding-path augmentation, various decoder paths are generated which predict HOI triplets in permuted sequences. Then, consistency regularization is applied across paths to enforce the predictions to be consistent. Parameter sharing and cross-matching were introduced as well to enhance learning. 

Our method is conceptually simple, and can be applied to a wide range of transformer architectures. Also, it does not require additional model capacity nor inference time. The substantial improvements on V-COCO and HICO-DET support our method's efficacy in various HOI detection tasks. Through further empirical studies, its capabilities to improve generalization and to encourage consistent representations are approved. 
\paragraph{Acknowledgements}
This work was partly supported by Institute of Information \& communications Technology Planning \& Evaluation (IITP) grant funded by the Korea government (MSIT) (No.2021-0-02312, Efficient Meta-learning Based Training Method and Multipurpose Multi-modal Artificial Neural Networks for Drone AI), (IITP-2022-2020-0-01819, the ICT Creative Consilience program); ETRI grant (22ZS1200, Fundamental Technology Research for Human-Centric Autonomous Intelligent System); and KakaoBrain corporation.
{\small
\bibliographystyle{ieee_fullname}
\bibliography{egbib}
}

\end{document}